\documentclass[conference]{IEEEtran}
\IEEEoverridecommandlockouts
\usepackage{cite}
\usepackage{amsmath,amssymb,amsfonts}
\usepackage{graphicx,subfig}
\usepackage{textcomp}
\usepackage{xcolor}
\usepackage{csquotes}

\usepackage{caption}
\usepackage{booktabs}
\usepackage{multirow}

\usepackage[ruled, lined, longend, linesnumbered]{algorithm2e}
\def\BibTeX{{\rm B\kern-.05em{\sc i\kern-.025em b}\kern-.08em
    T\kern-.1667em\lower.7ex\hbox{E}\kern-.125emX}}

\begin{document}

\title{Effective Scheduling Function Design in SDN through Deep Reinforcement Learning}

\author{\IEEEauthorblockN{Victoria Huang, Gang Chen, Qiang Fu}
\IEEEauthorblockA{\textit{School of Engineering and Computer Science} \\
\textit{Victoria University of Wellington}\\
Wellington, New Zealand \\
guiying.huang, aaron.chen, qiang.fu@ecs.vuw.ac.nz}
}

\maketitle

\begin{abstract}
Recent research on Software-Defined Networking (SDN) strongly promotes the adoption of distributed controller architectures. To achieve high network performance, designing a scheduling function (SF) to properly dispatch requests from each switch to suitable controllers becomes critical. However, existing literature tends to design the SF targeted at specific network settings. In this paper, a reinforcement-learning-based (RL) approach is proposed with the aim to automatically learn a general, effective, and efficient SF. In particular, a new dispatching system is introduced in which the SF is represented as a neural network that determines the priority of each controller. Based on the priorities, a controller is selected using our proposed probability selection scheme to balance the trade-off between exploration and exploitation during learning. In order to train a general SF, we first formulate the scheduling function design problem as an RL problem. Then a new training approach is developed based on a state-of-the-art deep RL algorithm. Our simulation results show that our RL approach can rapidly design (or learn) SFs with optimal performance. Apart from that, the trained SF can generalize well and outperforms commonly used scheduling heuristics under various network settings.
\end{abstract}


\section{Introduction} \label{sec:introduction}
Software-Defined Networking (SDN), a newly-emerged network paradigm, is notable for decoupling the control logic from the data forwarding function and forming a logically centralized control plane. With the separation of control and data planes, SDN greatly simplifies network management and enables efficient network configuration. To improve the scalability of SDN, distributed controller architectures \cite{onos,onix} have become a notable invention where multiple controllers are jointly deployed for scalable request processing. 

Apparently, high network performance depends on effective utilization of controller resources. This can be achieved by properly dispatching requests originated from every switch to suitable controllers chosen by a scheduling function (SF). Obviously, SF plays a vital role in the overall network performance. Motivated by this understanding, we aim to address the scheduling function design (SFD) problem in this paper. 

Particularly, the designed SF must satisfy both \emph{the time efficiency requirement ($R_1$)} and \emph{the generalization requirement ($R_2$)}.
In view of the fact that request dispatching must be performed in real time with minimum delay, the designed SF needs to be sufficiently efficient in practice. 
Moreover, SDN networks can vary significantly in the number and capacities of controllers. Thus, the designed SF should perform consistently well over different network settings.

Existing studies have considered either manual or automated design of similar functions for scheduling and resource allocation \cite{huang2017blac,nguyen2014automatic,park2018investigating}. Specifically, manually designed SFs such as weighted round-robin and first-come first-serve have been widely used in operating systems and cloud computing \cite{salot2013survey}. Obviously, the process of designing useful SFs is \emph{time-consuming} and requires \emph{a high level of domain expertise}. To address this difficulty, evolutionary computation (EC) techniques have been proposed to automatically design SFs for standard job shop scheduling problems \cite{nguyen2014automatic,park2018investigating}.
However, \emph{the evaluation process in EC is time-consuming and costly} since numerous randomly generated SFs must be extensively evaluated in either simulated or real-world environments.

Due to the above limitations, a new learning approach is highly desirable for our SFD problem. Recently, reinforcement learning (RL) has been successfully applied to various resource management problems \cite{mao2016resource,tesauro2006hybrid} and is considered to be a powerful paradigm for designing SFs with several key advantages. First, \emph{no domain knowledge of the environment is required}. RL can automatically learn the optimal solution while interacting with the unknown dynamic environment through a trial-and-error process. Second, 
\emph{RL can design new SFs mainly based on experiences/data obtained from an old SF} through a technique known as experience replay \cite{mnih2016asynchronous}. Thus, in comparison to an EC approach, the cost of training any new SFs can be greatly reduced.
Third, \emph{the scheduling problem under a specific network setting can be naturally formulated as a Markov Decision Process (MDP)} (detailed discussion can be found in Section \ref{subsec:mdp_mapping}), aiming to find an optimal policy using existing RL algorithms. In particular, a policy is a mapping from network states to a dispatching decision. 

Despite the clear advantages offered by RL, several major issues must be addressed. 
(1) The representation of a general policy remains a challenge. Typically, a policy can be directly represented as a neural network (NN) with fixed numbers of output nodes and each node represents one particular controller in the network. Such a representation apparently violates \emph{$R_2$} since the same policy is expected to function effectively in networks with different numbers of controllers. Moreover, as the number of controllers increases, the NN inevitably increases its complexity, which leads to long computational time to make a scheduling decision, potentially violating \emph{$R_1$} too. 
(2) It is difficult to maintain a good balance between exploration and exploitation for effective RL. Particularly, when multiple controllers are deployed, we expect to select each controller with a certain probability instead of deterministically choosing one controller. In such a way we can explore and learn the benefit of using each controller. However, this could easily cause performance degradation without carefully controlling the level of exploration. 
As far as we know, none of the existing works have considered and solved the above issues. 

In this paper, a new RL-based SFD method is proposed 
with the aim to automatically learn an effective and efficient SF for general use. 
The following contributions have been achieved.
(1) Instead of representing the SF as an RL policy, a new dispatching system is proposed as a practical implementation of an RL policy, in which the SF is represented as an NN taking the states of each individual controller as input and outputting its \textquote{priority}.
(2) Given the priorities of all controllers, a probability selection scheme is proposed as part of the dispatching system to cope with the exploration-exploitation dilemma. With the proposed selection scheme, only controllers with high priorities have the possibilities to be selected. As a consequence, the long-term network performance can be improved without sacrificing exploration whenever a deterministic controller selection scheme is adopted.
(3) Along with the new dispatching system, a new training system is developed based on a state-of-the-art actor-critic RL algorithm \cite{schulman2017proximal}. In particular, a new gradient calculation technique is derived for learning the SF. Apart from that, a new training scheme is proposed so as to constantly and adaptively improve the performance of the SF under a variety of network settings.

\section{Related Work}
In recent years, distributed controller architectures \cite{onos,onix} have been widely adopted in SDN to enhance the network performance. Although multiple controllers can be deployed in the control plane, the network performance still heavily relies on effective utilization of the controller resources. Thus, designing an effective and efficient SF for request dispatching is of great importance.

In the literature, there exist SFs in the form of heuristics designed by human experts. For example, a weighted round-robin heuristic is designed to proportionally forward requests to controllers based on their processing capacities. BLAC \cite{huang2017blac} randomly sampled a small number of controllers and sent requests to the least loaded one. Similar approaches can also be found in literature \cite{elasticon,wang2018effective}.
Although such manually designed SFs are intuitive and simple in nature, the design process is time-consuming and requires substantial domain knowledge. Moreover, the performance of manually designed SFs could also vary significantly, depending on specific network settings. 

To address these limitations, EC techniques have been widely applied for automatically designing SFs. For instance, Su et al. \cite{nguyen2014automatic} proposed a multi-objective Genetic Programming (GP) approach for handling dynamic job shop scheduling problems. 
Although promising results have been obtained in the literature, these SFs are generally designed offline and cannot easily and quickly adapt to the never-ending changes in the network environment \cite{branke2016automated}. Besides, each newly evolved SF must be extensively tested in either simulated or real-world environments, which is time-consuming and costly.  


Recently, a completely different design approach based on RL has been studied in the literature \cite{mao2016resource,tesauro2006hybrid,chinchali2018cellular}. Tesauro et al. \cite{tesauro2006hybrid} proposed an RL-based approach to automatically allocate the server resources in data centers. DeepRM \cite{mao2016resource} tackled the multi-resource cluster scheduling problem using policy search to optimize various objectives, e.g., average job completion time and resource utilization. Chinchali et al. \cite{chinchali2018cellular} leveraged the delay-tolerant feature of IoT traffic and developed an RL-based scheduler to handle traffic variation so that the network utilization can be constantly optimized. 

All these RL-based methods assume that the policy is represented by an NN. The dimension of the outputs is fixed and essentially equal to the number of controllers in an SDN network. However, when the network environment changes, e.g., more controllers are added due to the traffic growth, the trained policy is no longer applicable. Therefore, existing RL methods cannot be directly utilized to solve our SFD problem.    

In view of above reasons, a new RL-based approach is proposed. 
According to our discussion in Section \ref{sec:introduction}, our approach satisfies both \emph{$R_1$} and \emph{$R_2$} and has the key strength of leveraging RL to effectively schedule requests within the SDN network so as to optimize the network performance.

\section{Understanding the SFD Problem} \label{sec:understanding_scheduling}
In this section, we will introduce the SFD problem by discussing the key concepts and modeling the problem as an MDP that lays the foundation of an RL-based solution. 

\subsection{The SFD Problem in SDN} \label{subsec:the SFD problem in SDN}
Before introducing the SFD problem, we first introduce the network environment where the SF is applied. To ease discussion, let us consider an SDN network composed of $N_c$ controllers and $N_s$ switches. Specifically, the processing capacities of the $N_c$ controllers can be captured through $\boldsymbol{\alpha} = [\alpha_1, ..., \alpha_{N_c}]$ where $\alpha_{n_c} (n_c = 1, ..., N_c)$ represents the maximum number of requests that controller $C_{n_c}$ can process within a second. Packets arrive at switches constantly in the data plane. Note that when a new packet arrives at a switch, the switch will generate a request and pass it to a controller for processing. The packet arrival rate, therefore, is identical to the rate at which requests are generated by switches.
We assume that packets arrive randomly at switches with respective arrival rates $\boldsymbol{\lambda}=[\lambda_1, ..., \lambda_{N_s}]$.
Similar to existing works \cite{huang2017blac,wang2016dynamic}, we assume that the time for processing each request by the same controller is roughly identical. However, it should also be noted that since controllers have different capacities, the processing time would change from one controller to another.

Once a request is generated at a switch, it will be immediately forwarded to a controller for processing with the help of our dispatching system. Since requests are generated (and dispatched) at different time, the dispatching of each request is considered as a separate time step of our dispatching system. This assumption can be flexibly supported with different implementation of the dispatching system. For example, we can install a separate and identical dispatching system on each switch to handle its request. 
The communication delay between the switches and controllers can be described by a matrix $\boldsymbol{D}$, where each element $D_{n_s}^{n_c}$ represents the delay between switch $S_{n_s}$ and controller $C_{n_c}$. 

After processing any request, the controller will send a response back to the corresponding switch. The time interval measured by the switch between sending a request and receiving the response is defined as the request response time $\tau$. Apart from request processing, controllers will periodically report their status $\boldsymbol u = [u_1, ..., u_{N_c}]$ in terms of current resource utilization to all switches. Without loss of generality, we also assume that each controller maintains a request queue and processes requests in an FIFO manner \cite{huang2017blac}.

With a properly designed SF, we expect to reduce the average request response time. To achieve this, \emph{$R_1$}
is crucial to avoid potential network performance degradation. 
We consider that the NN can meet \emph{$R_1$} because small feed-forward NNs can be quickly processed with the support of efficient and performance-optimized software (e.g., TensorFlow \cite{Tensorflow}) and dedicated processing chips \cite{tpu}. 
In addition to \emph{$R_1$}, a carefully designed SF must generalize well (i.e., \emph{$R_2$}). Since the cost of evaluation or training an SF in a production network can be high, the SF needs to be evaluated or trained in advanced. 
Note that different SDN networks can vary significantly in terms of number and capacities of controllers. Even within the same network, the number of controllers may change dynamically to accommodate the traffic fluctuation. Thus, a generally applicable SF should be able to immediately cope with these variations. To achieve this goal, we need to first model an SFD problem as an MDP.

\subsection{Modeling the SFD Problem as an MDP} \label{subsec:mdp_mapping}
An MDP is usually described by a 4-tuple $(\mathcal{S}, \mathcal{A}, \mathcal{P}, \mathcal{R})$. At each time step $t$, an agent observes its current state $\boldsymbol s_t \in \mathcal{S}$ while interacting with an unknown environment and takes an action $a_t \in \mathcal{A}$ chosen from a policy $\pi_{\boldsymbol \theta}$. A policy $\pi_{\boldsymbol \theta}$ is often considered as a parametric function of $\boldsymbol \theta$, which maps $\mathcal S$ to a probability distribution over $\mathcal A$. After performing $a_t$, the agent receives a reward given by the reward function $\mathcal R(\boldsymbol{s}_t, a_t)$ and enters the next state $\boldsymbol s_{t+1}$ decided by the state transition probabilities $\mathcal P(\boldsymbol{s}_{t+1}|\boldsymbol{s}_t, a_t)$. The goal for RL in a finite horizon $T$ is to learn the policy $\pi_{\boldsymbol \theta}$ so as to maximize the expected long-term cumulative reward defined below:
\begin{equation}
\label{sec3_eqt:rl_objective}
V_{\pi_{\boldsymbol \theta}}(\boldsymbol{s}) = E\left \{\sum_{t=0}^{T} \mathcal R(\boldsymbol{s}_t, a_t)|\boldsymbol{s}_0=\boldsymbol{s}, \pi_{\boldsymbol{\theta}} \right \} 
\end{equation}

The SFD problem can be naturally formulated as an MDP. Specifically, the network reaches a new time step $t$ is defined whenever one new request is generated by a switch in the data plane. At each time step $t$, $\boldsymbol s_t$ contains all current and historical network information, such as $\boldsymbol \lambda$ and $\boldsymbol u$; $a_t$ is the controller selected by the policy $\pi_{\boldsymbol \theta}$ and $\pi_{\boldsymbol \theta}$ is the dispatching system (Fig.\ref{sec4_fig:policy}) to be introduced in Section \ref{sec:proposed_RL_sys}. After dispatching the requests to the chosen controller, the network keeps operating until the next request is generated. At that moment the network enters the next state $\boldsymbol s_{t+1}$. In order to train $\pi_{\boldsymbol \theta}$ towards optimizing our objective (i.e., minimizing the average request response time), we define the reward as
\begin{equation}
\mathcal R(\boldsymbol s_t,a_t)=\sum_{j \in \mathcal J_t} \frac{1}{\tau_{j}}
\end{equation}
where $\tau_{j}$ is the response time of request $r_j$ and $\mathcal J_t$ stands for the set of requests, for which the corresponding response from controllers have been received by the respective switches in between two consecutive time steps $t$ and $t+1$. 



Clearly, each request $r_j$ contributes $\frac{1}{\tau_j}$ to the total reward. Guided by this reward, an RL algorithm is strongly motivated to receive more responses from controllers and to reduce the average response time simultaneously. By maximizing the cumulative reward in (\ref{sec3_eqt:sch_objective}), we can therefore fulfill our goal of reducing the request response time and improving the network performance.
\begin{equation}
\label{sec3_eqt:sch_objective}
\underset{\boldsymbol \theta}{\text{max}} \; V_{\pi_{\boldsymbol \theta}}(\boldsymbol{s}) 
 =\underset{\boldsymbol \theta}{\text{max}} \; E\left \{\sum_{t=0}^{T} \sum_{j \in J_t} \frac{1}{\tau_{j}} |\boldsymbol{s}_0=\boldsymbol{s}, \pi_{\boldsymbol \theta} \right \}
\end{equation}

\section{Proposed RL-based Approach for SFD} \label{sec:proposed_RL_sys}
\begin{figure}[!tbp]
\centering
\includegraphics[width=1.0\linewidth]{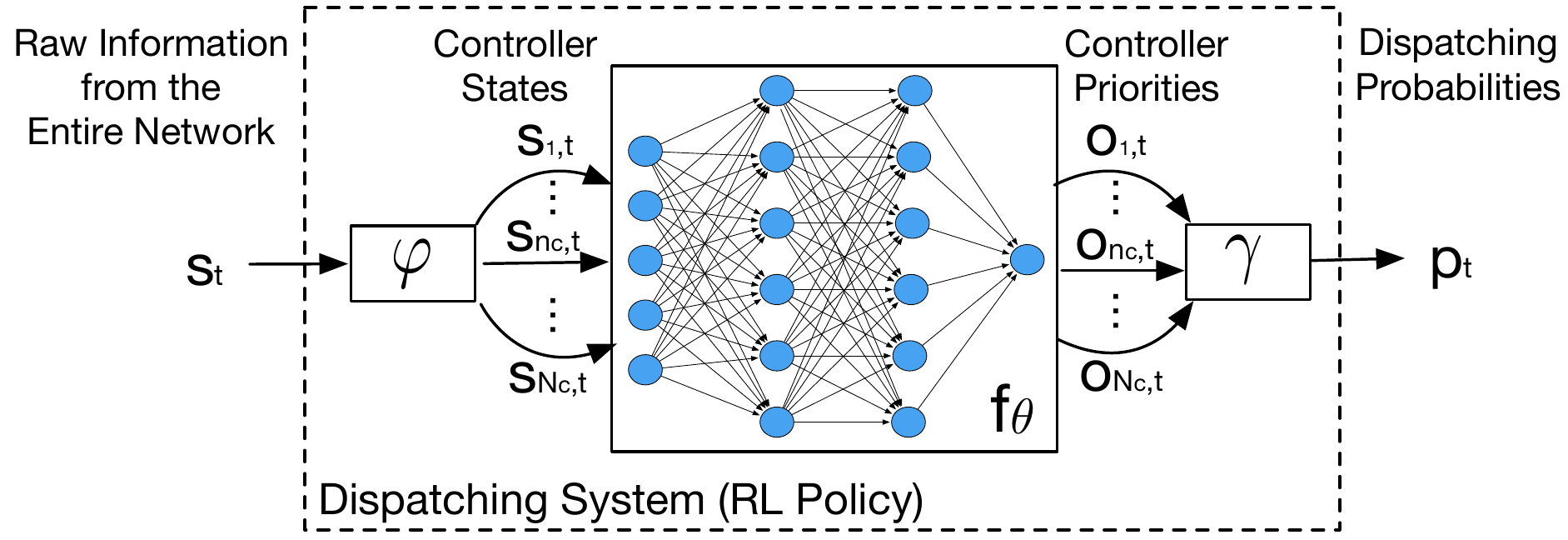}
\caption{Dispatching System Design.}\label{sec4_fig:policy}
\vspace{-10pt}
\end{figure}

Our proposed RL-based approach for SFD consists of two major systems: the dispatching system in Fig. \ref{sec4_fig:policy} and the training system in Fig. \ref{sec4_fig:training_system}. In particular, the dispatching system, which is also known as the RL policy, chooses a suitable controller for each incoming packet so as to minimize the response time. On the other hand, the training system is in charge of optimizing the SF under varied network settings for general use.

\subsection{Dispatching System Design} \label{subsec:dispatching_system}
By modeling the SFD problem as an MDP, RL algorithms can be utilized to design a general, efficient, and effective SF. However, as we mentioned in Section \ref{sec:introduction}, existing policy representation fails to meet both \emph{$R_1$} and \emph{$R_2$}. Thus, a new policy representation  must be designed.
Inspired by the successful use of dispatching rules for supporting diverse job shop scheduling tasks \cite{nguyen2014automatic,park2018investigating}, we decide to employ an SF to determine the priority for each controller to process every incoming packet. Consequently, the same SF is capable of dispatching requests in a network with arbitrary number of controllers. Moreover, we adopt NN in this paper to improve the expressiveness and trainability of the SF. 
It is important to note that in job shop scheduling problems, the job with the highest priority will always be processed first according to the dispatching rule. In comparison, we choose to give more controllers non-negligible opportunity of processing a request, thereby encouraging the exploration during the training process in hope of achieving higher performance.


In line with this idea, a new dispatching system has been designed in Fig. \ref{sec4_fig:policy}. The system consists of three modules - the state extractor $\varphi$, the SF $f_{\boldsymbol \theta}$, and the probability projector $\gamma$. In particular, when a new request is generated at a switch at time $t$, the dispatching system will select a controller using the following steps: First of all, $\varphi$ extracts the specific state information $\boldsymbol s_{n_c,t}$ for each controller from the raw entire network information $\boldsymbol s_t$. Then the state information $\boldsymbol s_{n_c,t}$ of each controller will be processed individually and sequentially by the SF $f_{\boldsymbol \theta}$ which outputs a corresponding priority value $o_{n_c,t}$. Given all controllers' priorities $\boldsymbol o_{t} = [o_{1,t}, ..., o_{N_c,t}]$, $\gamma$ is activated to map $\boldsymbol o_{t}$ into dispatching probabilities $\boldsymbol p_{t}$. Based on $\boldsymbol p_{t}$, a controller is selected to process the new request.


Obviously, the output priority $o_{n_c,t}$ of controller $C_{n_c}$ depends heavily on its state input $\boldsymbol s_{n_c,t}$ as shown in Fig. \ref{sec4_fig:policy}. Thus, selecting suitable state information for each controller is critical.
Intuitively, the preference of choosing controller $C_{n_c}$ relies on its history. If its response time $\tau_{n_c}$ or utilization $u_{n_c}$ dramatically increased recently, the controller is very likely to be heavily loaded in the near future. Sending requests to that controller is likely to result in long response time. Similarly, the request arrival rate $\lambda_{n_s,t}$ at switch $S_{n_s}$ at time step $t$ should also be included because the controller's future utilization is directly affected by the number of requests originated from switch $S_{n_s}$. Apart from that, the preference of choosing $C_{n_c}$ also depends on its processing capacity $\alpha_{n_c}$ and communication delay $D^{n_c}_{n_s}$. Therefore, the information mentioned above should all be included in $C_{n_c}$'s state information $\boldsymbol s_{n_c,t}$.


Given the controller state $\boldsymbol s_{n_c,t}$, $f_{\boldsymbol\theta}$ computes the controller's priority $o_{n_c,t}$ through an NN parameterized by $\boldsymbol \theta$ which will be optimized using an adapted RL algorithm elaborated in Section \ref{subsec:ppo} and \ref{subsec:training_system}. The calculated priorities of all controllers $\boldsymbol{o}_t = [o_{1,t},...,o_{N_c,t}]$ are then forwarded to the next system module. 

After receiving the priorities $\boldsymbol o_{t}$ from all controllers, the probability projector $\gamma$ maps the priorities into dispatching probabilities $\boldsymbol{p}_t = [p_{1,t}, ..., p_{N_c,t}]$. One of the most widely used mappings is the softmax function \cite{sutton1998introduction}. However, we do not consider it as an appropriate option because a non-zero probability will always be assigned to a controller even though the controller is clearly not a suitable candidate for processing the pending request (e.g., the controller is too far away from a switch or has very low capacity). This will inevitably result in network performance degradation. On the other hand, a deterministic approach which always selects the controller with the highest priority is also inappropriate. 
Because purely exploiting the currently best controller prevents an RL algorithm from exploring other suitable controller candidates that can bring more benefits in reducing the average response time.

To balance the trade-off between exploration and exploitation, an Euclidean projection method \cite{wang2013projection} is adopted here. In particular, we define $\boldsymbol {\Tilde{o}}_{t}=[\Tilde{o}_{1,t}, ..., \Tilde{o}_{N_c,t}]$ as the normalized priorities of $\boldsymbol o_{t}$ and is sorted in descending order, where $\Tilde{o}_{i,t}$ represents the normalized priority of the controller with the $i^{th}$ highest priority. Note that $\boldsymbol o_{t}$ is the output of the NN which is guaranteed to be non-negative by using a softplus activation function in the output layer.
We expect to compute the Euclidean projection of $\boldsymbol {\Tilde{o}}_{t}$ to $\boldsymbol p_t$ so that the Euclidean distance between $\boldsymbol {\Tilde{o}}_{t}$ and $\boldsymbol p_t$ can be minimized.


Existing studies \cite{wang2013projection} showed that the Euclidean projection problem can be solved 
by assigning non-zero probabilities to the $m$ controllers with the highest priorities while setting $0$ probabilities to the remaining controllers. The exact solution to the Euclidean projection problem is further given as follows:
\begin{equation} \label{sec4_eqt:g}
p_{i,t} = 
\begin{cases}
\Tilde{o}_{i,t}+\frac{1}{m}(1-\sum_{j=1}^{m} \Tilde{o}_{j,t}),  \; \text{if} \;1 \leq i \leq m \\
{}0, \; \text{otherwise}
\end{cases}
\end{equation}
where $m$ is a hyper-parameter for our dispatching system.

With \eqref{sec4_eqt:g}, the requests will be always dispatched to \textquote{appropriate} controllers (i.e., controllers with high priorities) so as to improve the long-term network performance. Furthermore, by manipulating the value of $m$, we can explicitly control the level of exploration.

\subsection{Adapting PPO to Train the SF} \label{subsec:ppo}
Among existing RL algorithms, proximal policy optimization (PPO) is selected for training the SF because of several reasons. First of all, PPO can perform multiple epochs of minibatch policy update using previously sampled data, greatly improving sample efficiency. Secondly, PPO employs only first-order optimization which is more computationally efficient compared to other RL algorithms \cite{trpo}. Finally, PPO has been widely and successfully used in many problem domains. Studies \cite{schulman2017proximal} have shown that PPO can outperform many state-of-the-art algorithms such as TRPO \cite{trpo} and A2C \cite{mnih2016asynchronous} on many difficult RL problems. It is particularly effective at training functions modeled as deep NNs. 
While we only use PPO in this paper, our research does not rule out the possibilities of using other RL algorithms.

As a prominent and highly efficient actor-critic algorithm, PPO uses an NN to approximate the value function which is then used to train the SF. In order to apply PPO, 
an NN denoted as $f_{\boldsymbol \omega}$ parameterized by $\boldsymbol \omega$ representing the value function is required.
In line with \emph{$R_2$}, the number of inputs to $f_{\boldsymbol \omega}$ should not change even when the network setting alters. In our simulation studies, we found it useful to feed $f_{\boldsymbol \omega}$ with high-level statistics $\boldsymbol s'_t$ that accurately capture the performance and operation of our SDN network in the recent past. 
Thus, in our experimental study, the inputs $\boldsymbol s'_t$ for $f_\omega$ contain the total control plane capacity, the weighted average communication delay, the overall request arrival rate, and the recent history of both average response time and the control plane utilization. 

Given the value function $f_{\boldsymbol \omega}$, PPO obtains the optimal policy $\pi_{\boldsymbol \theta}$ by maximizing the following clipping function:
\begin{equation} \label{sec4_eqt:ppo}
C = \underset{\boldsymbol \theta}{\text{max}} \; E\left \{ \text{min} (r_t(\boldsymbol \theta)A_t,
\text{clip}(r_t(\boldsymbol \theta),
1-\epsilon, 1+\epsilon) A_t) \right \}
\end{equation} 
where  $\epsilon$ is a hyper-parameter (e.g., $0.2$), $A_t$ is the advantage function that can be estimated through $f_{\boldsymbol \omega}$ \cite{schulman2015high}, and $r_t(\boldsymbol \theta)=\frac{\pi_{\boldsymbol \theta}(a_t|\boldsymbol s_t)}{\pi_{\boldsymbol \theta_{old}}(a_t|\boldsymbol s_t)}$. It is straightforward to see that the policy $\pi_{\boldsymbol \theta}$ can be improved by repeatedly updating the policy parameters $\boldsymbol \theta$ along the direction of $\frac{\partial C}{\partial \boldsymbol \theta}$.

In particular, for any state $\boldsymbol s_t$, the gradient $\frac{\partial C}{\partial \boldsymbol \theta}$ can be calculated as below:
\begin{equation} \label{sec4_eqt:gradient}
\frac{\partial C}{\partial \boldsymbol \theta} = 
\begin{cases}
0, \;  \text{if}\;\left \{ \begin{matrix} r_t < 1- \epsilon \\ A_t < 0 \qquad \end{matrix} \right. \text{or} \;\left \{ \begin{matrix} r_t > 1+ \epsilon \\ A_t > 0 \qquad \end{matrix} \right. \text{or} \; \pi_{\boldsymbol \theta}= 0 \\
{}\frac{A_t}{\pi_{\boldsymbol \theta_{\text{old}}}} \frac{\partial \pi_{\boldsymbol \theta}}{\partial \boldsymbol \theta}, \; \text{otherwise}
\end{cases}
\end{equation}

Note that $\pi_{\boldsymbol \theta}$ in Fig. \ref{sec4_fig:policy} combines the mapping function $\gamma$ described in \eqref{sec4_eqt:g} and the NN $f_{\boldsymbol \theta}$.
Therefore, calculating $\frac{\partial C}{\partial \boldsymbol \theta}$ requires extra effort.
A new technique is developed in this paper to calculate $\frac{\partial C}{\partial \boldsymbol \theta}$ which will be illustrated by the following example.

\textbf{Gradient Calculation Example:} 
In this example, we consider a network with $3$ controllers and set the hyper-parameter $m$ in \eqref{sec4_eqt:g} to be $2$. Assume that the $3$ controllers' priorities at time step $t$ is $\boldsymbol{o}_t=[o_{1,t}, o_{2,t}, o_{3,t}]$ and $o_{1,t} \geq o_{2,t} \geq o_{3,t}$. Then the sorted and normalized priorities $\boldsymbol {\Tilde{o}}_{t}$ can be represented as 
\begin{equation}
\boldsymbol {\Tilde{o}}_{t}=[\frac{o_{1,t}}{\sum_{j=1}^3 o_{j,t}}, \frac{o_{2,t}}{\sum_{j=1}^3 o_{j,t}}, \frac{o_{3,t}}{\sum_{j=1}^3 o_{j,t}}]
\end{equation}
The dispatching probabilities in \eqref{sec4_eqt:g} can be determined as:
\begin{equation} 
p_{i,t} = 
\begin{cases}
\Tilde{o}_{i,t}+0.5 \times \Tilde{o}_{3,t} = \frac{o_{i,t}+0.5 \times o_{3,t}}{\sum_{j=1}^3 o_{j,t}},  \; i = 1, 2 \\ 
{} 0, \; i = 3
\end{cases}
\end{equation}
Following the chain rule, the gradient of $\pi_{\boldsymbol{\theta}}$ can be calculated as below if the controller with the $i^{th}$ highest priority is selected:
\begin{equation}
\text{\footnotesize$
\begin{split}
& \frac{\partial \pi_{\boldsymbol \theta}(\boldsymbol s_t, a_t)}{\partial \boldsymbol \theta} = 
\frac{\partial p_{i,t}}{\partial {o_{1,t}}}\frac{\partial {o_{1,t}}}{\partial {\boldsymbol {\theta}}} 
+ \frac{\partial p_{i,t}}{\partial {o_{2,t}}}\frac{\partial {o_{2,t}}}{\partial {\boldsymbol {\theta}}}
+\frac{\partial p_{i,t}}{\partial {o_{3,t}}}\frac{\partial {o_{3,t}}}{\partial {\boldsymbol {\theta}}}\\ 
& = \frac{\frac{\partial {o_{i,t}}}{\partial{\boldsymbol \theta}}+0.5  \frac{\partial {o_{3,t}}}{\partial{\boldsymbol \theta}}}{\sum_{j=1}^3 o_{j,t}} - 
\frac{(o_{i,t}+0.5 o_{3,t})(\frac{\partial {o_{1,t}}}{\partial{\boldsymbol \theta}}+\frac{\partial {o_{2,t}}}{\partial{\boldsymbol \theta}}+\frac{\partial {o_{3,t}}}{\partial{\boldsymbol \theta}})}{(\sum_{j=1}^3 o_{j,t})^2}
\end{split}
$}
\end{equation}
where $\frac{\partial {o_{n_c,t}}}{\partial{\boldsymbol \theta}} =
\frac{\partial{f_{\boldsymbol{\theta}}}}{\partial{\boldsymbol \theta}}$ is the gradient of the SF $f_{\boldsymbol \theta}$ given the controller state $\boldsymbol s_{n_c,t}$. 

Note that this new technique for calculating the derivative can be easily extended to the case with arbitrary number of controllers. With the help of TensorFlow \cite{Tensorflow}, the derivative calculation can also be fully automated in our training system, regardless of how many controllers are involved.

\begin{figure}[!tbp]
\centering
\includegraphics[width=0.85\linewidth]{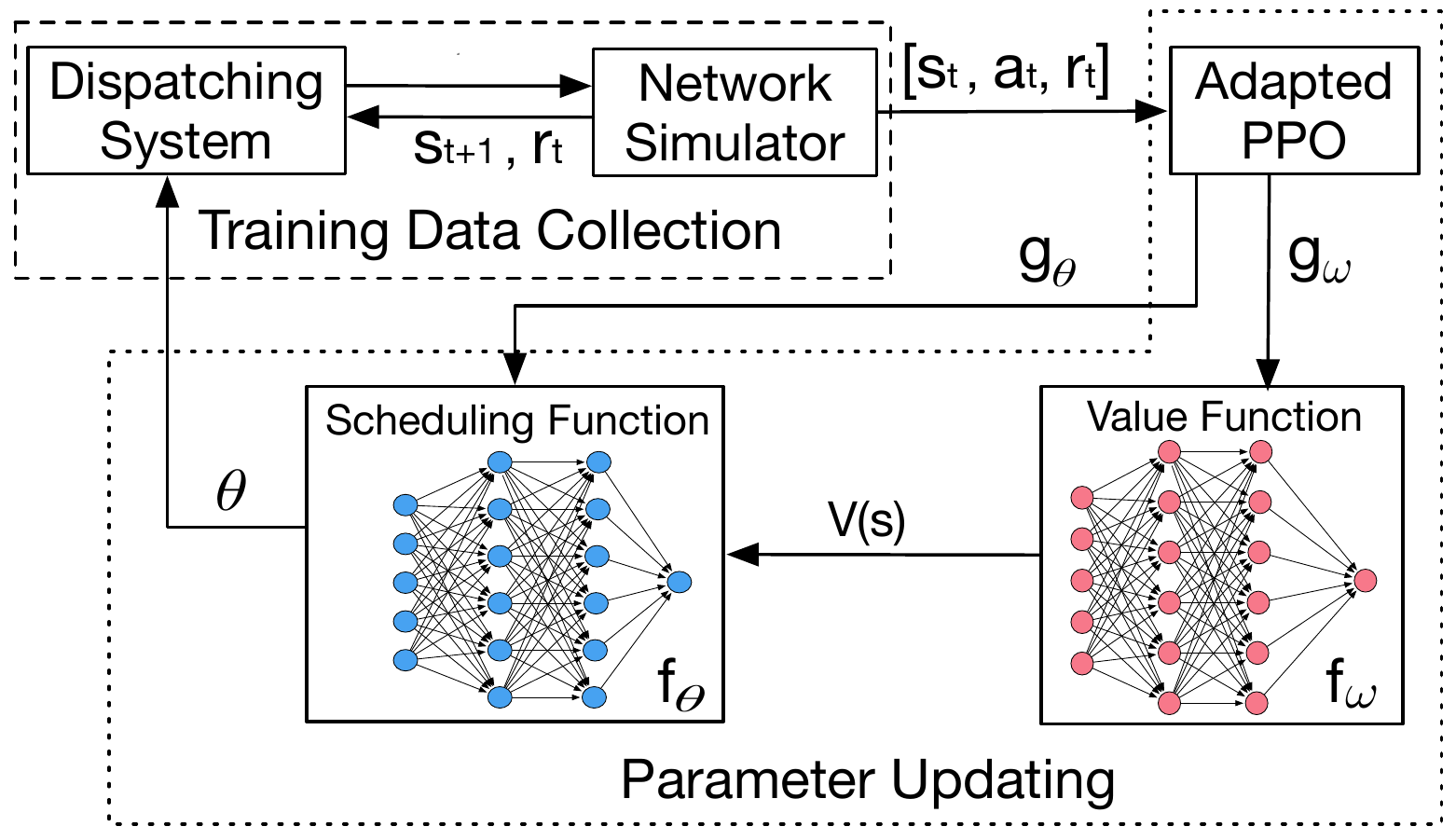}
\caption{Training System Design.}
\label{sec4_fig:training_system}
\vspace{-10pt}
\end{figure}

\subsection{Training System Design} \label{subsec:training_system}
In this section, we will discuss how to simultaneously train both the SF $f_{\boldsymbol \theta}$ and the value function $f_{\boldsymbol \omega}$.

\setlength{\textfloatsep}{9pt}
\begin{algorithm}
  \caption{PPO-based Algorithm for SF Training.} \label{sec4_alg:training}
    \For{Each network setting $N_s=1:N_{set}$}{%
        \For{Each episode $N_e=1: N_{ep}$}{%
            \While{$t < t_{\text{max}}$}{
            Perform one learning iteration update:\\
            (1) Collect training data $(\boldsymbol s_t, \boldsymbol a_t, \boldsymbol r_t)$: Run dispatching system in network simulator for $n$ time steps\;
    		(2) Adapt PPO to compute gradients $\boldsymbol {g_w}$ and $\boldsymbol {g_\theta}$\;
    		(3) Update NN parameters: $\boldsymbol \theta_{old} \leftarrow \boldsymbol \theta $ and $\boldsymbol \omega_{old} \leftarrow \boldsymbol \omega$ 
            }
        }
    }

\end{algorithm}



To understand the training process, we first define a network setting as a 3-tuple $\textbf {Env} = (\boldsymbol \alpha, \boldsymbol \lambda, \boldsymbol D)$ including the controller capacities, packet arrival rates, and communication latencies. An episode is further defined as one network simulation based on a specific setting $\textbf {Env}$, which starts from an initial state where no packets have started to flow through the network and ends when the simulation time reaches a predefined value $t_{max}$. Simulation details will be provided in Section \ref{sec:simulation}.

As shown in Algorithm \ref{sec4_alg:training}, the training is performed in a sequential manner. In particular, $N_{set}$ network settings $[\textbf {Env}_1, ..., \textbf {Env}_{N_{set}}]$ will be simulated in sequence. For each $\textbf {Env}$, PPO is used to train the SF for $N_{ep}$ episodes. Within each episode, multiple learning iterations are performed. 

In particular, a learning iteration consists of $n$-time-step simulation.   
As shown in Fig. \ref{sec4_fig:training_system}, within each learning iteration, the dispatching system equipped with $f_{\boldsymbol \theta_{old}}$ is used for dispatching requests in the network simulator for $n$ time steps. After collecting the corresponding information including states, actions, and rewards from the simulated network, PPO is further activated to train the SF.

\section{Simulation} \label{sec:simulation}
This section reports the performance evaluation of the proposed RL-based approach for SFD. 

\textbf{Simulation Setting:} 
In our simulation, we adopt the same NN architecture given in PPO \cite{schulman2017proximal} for both the SF $f_{\boldsymbol{\theta}}$ and value function $f_{\boldsymbol{\omega}}$, which is a fully connected multilayer perceptron with two hidden layers of 64 units and tanh activation. Meanwhile, we set the hyper-parameters following the PPO Mujoco setting \cite{schulman2017proximal} and $m$ in \eqref{sec4_eqt:g} to be $2$.
Each network setting is trained for $2$ episodes and each episode is initialized with $0\%$ utilization for all controllers and $0$ packets in the network. 
The requests are generated following the Poisson distributions with predefined arrival rates. Each episode runs for $240$ seconds which is assumed to be sufficiently long for the network to enter and stay in a stationary condition. For each episode, we set $1024$ simulation steps as one learning iteration. 

During our simulation, the SF is trained with one switch since the learned SF will be individually deployed on each switch as we mentioned in \ref{subsec:the SFD problem in SDN} and training the SF with one switch can also reduce the training costs. Although the SF is trained with one switch, it can effectively work in the network with multiple switches deployed, which will be demonstrated in the testing section.

\begin{figure}[!tbp]
  \begin{center}
    \subfloat[Cumulative reward.]{\label{sec5_fig:2ctlreward}\includegraphics[width=.4\linewidth]{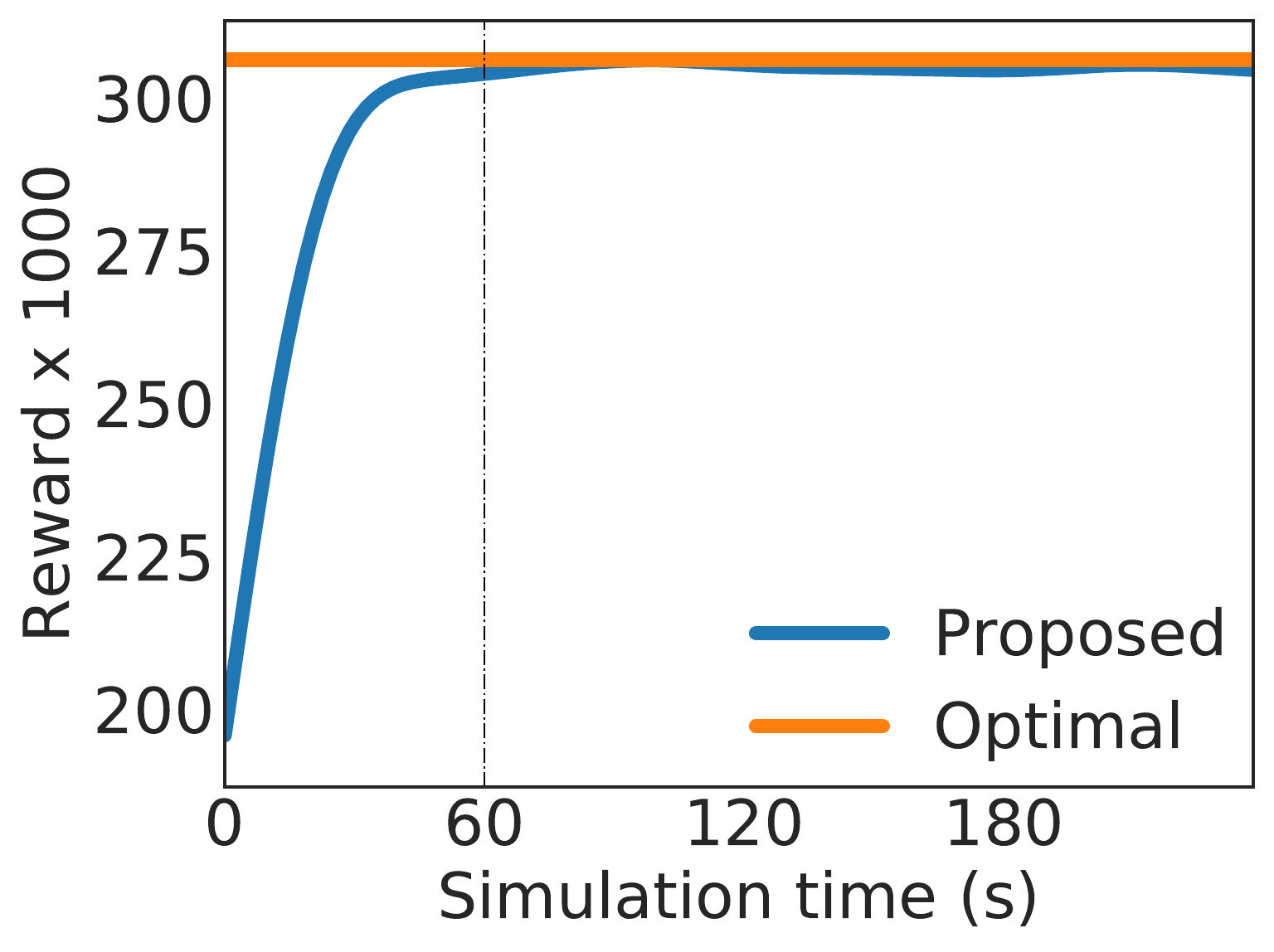}}
    \subfloat[Average response time.]{\label{sec5_fig:2ctlresptime}\includegraphics[width=.4\linewidth]{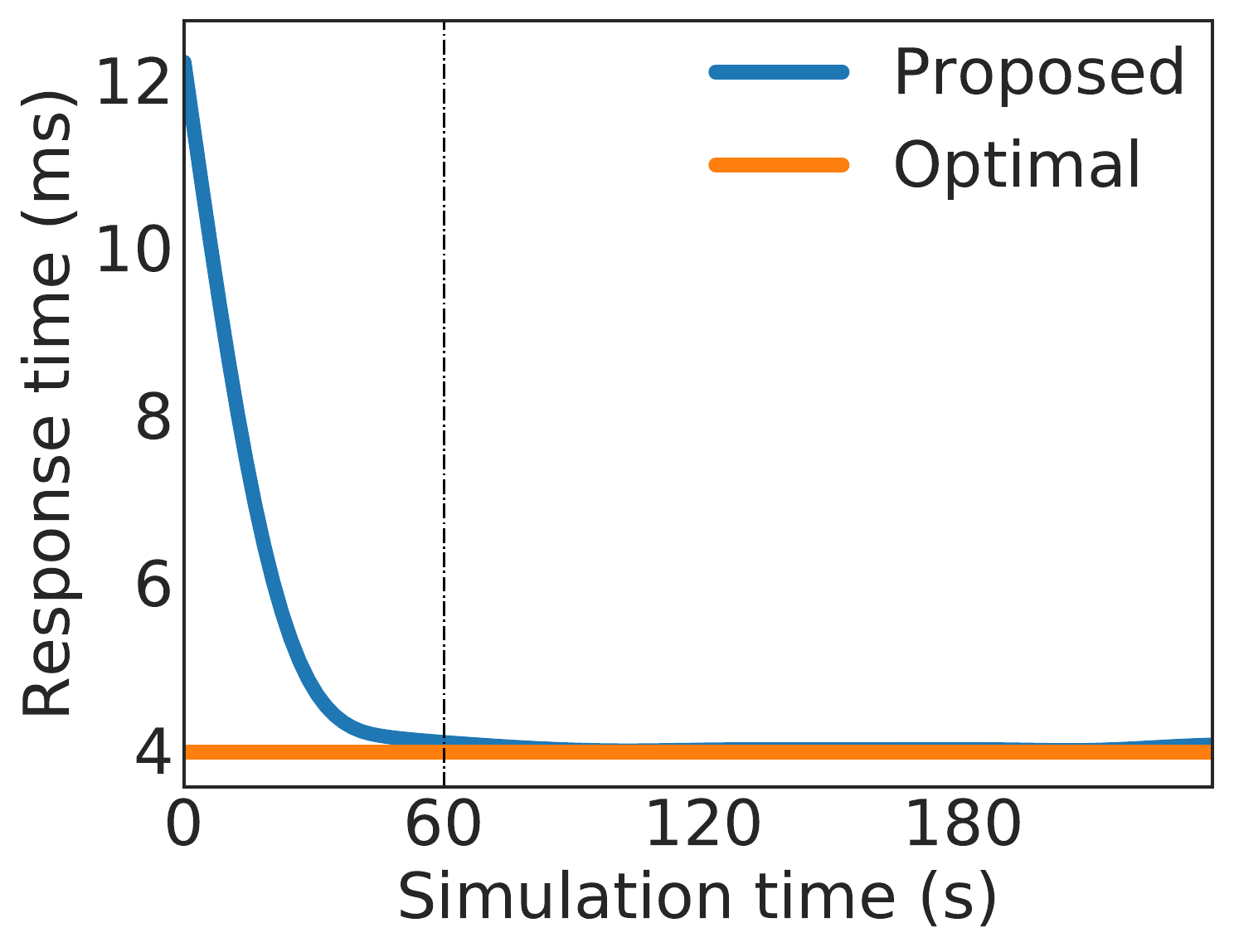}}
  \end{center}
  \vspace{-10pt}
  \caption{Training performance of our proposed SF on Env 1.}
  
\label{sec5_fig:training_performance}
\end{figure}

\textbf{Training Performance:}
We start with demonstrating the training performance when only one network setting (Env $1$ in Table \ref{sec5_tab:network_setting}) is utilized for training.  
The learning curves showing the total reward and average response time averaged across five runs of the algorithm are demonstrated in Fig. \ref{sec5_fig:training_performance}. 
It can be observed from Fig. \ref{sec5_fig:training_performance}(a) that, 
the cumulative reward grows quickly from around $200,000$ to above $300,000$ within $60$s of the corresponding simulated network operation time. Similarly, a sharp decrease in average response time can also be observed from Fig. \ref{sec5_fig:training_performance}(b) which indicates that the trained SF can rapidly converge to the optimal performance.


\begin{table}[!tbp]
\centering
\caption{Network settings for training and testing.}
\label{sec5_tab:network_setting}
\tiny
\resizebox{.43\textwidth}{!}{
\setlength\tabcolsep{1.5pt}
\begin{tabular}{c|c|c|c|c|c|c|c} \hline
\multicolumn{1}{c|}{\multirow{2}{*}{}} & \multicolumn{3}{c|}{Training} & \multicolumn{4}{c}{Testing} \\ \cline{2-8}
\multicolumn{1}{c|}{} & Env 1 & Env 2 & Env 3 & Env 4 & Env 5 & Env 6 & Env 7 \\ \hline 
\begin{tabular}[c]{@{}c@{}}Total request arrival\\ 
rate $\boldsymbol \lambda$ ($\times$1000pkt/s) \end{tabular} & 5 & 4 & 6 & 15 & 20 & 15 & 25 \\ \hline 
Num. controller $N_c$ & 2 & 2 & 2 & \multicolumn{2}{c|}{3} & \multicolumn{2}{c}{4}\\ \hline 
\begin{tabular}[c]{@{}c@{}}Processing capacities\\ $\boldsymbol \alpha$($\times$1000pkt/s)\end{tabular} & {[}9, 9{]} & {[}15, 6{]} & {[}9, 12{]} & \multicolumn{2}{c}{{[}6, 9, 12{]}} & \multicolumn{2}{|c}{{[}6, 9, 12, 15{]}} \\ \hline
\begin{tabular}[c]{@{}c@{}}Communication\\
delay $\boldsymbol D$(s) \end{tabular} & \begin{tabular}[c]{@{}c@{}}{[}0.002, \\ 0.02{]}\end{tabular} & \begin{tabular}[c]{@{}c@{}}{[}0.01, \\ 0.01{]} \end{tabular} & \begin{tabular}[c]{@{}c@{}}{[}0.005, \\ 0.04{]} \end{tabular} & \multicolumn{2}{c|}{\textbackslash} & \multicolumn{2}{c}{\textbackslash} \\ \hline 
\end{tabular}
}
\vspace{-4mm}
\end{table}

In order to train a generally applicable SF, we then perform the training over a series of different network settings as described in Section \ref{subsec:training_system}. In particular, $3$ network settings (Env $1$ - Env $3$ shown in Table \ref{sec5_tab:network_setting}) are used during the training process. 
The corresponding training performance is similar to Fig. \ref{sec5_fig:training_performance}, which is excluded here due to the space limit. 

\begin{figure}[!tbp]
  \begin{center}
    \subfloat[South American Network with arrival rate $15000$pkt/s (Env 4).]{\label{sec5_fig:4ctl_3sw}\includegraphics[width=.46\linewidth]{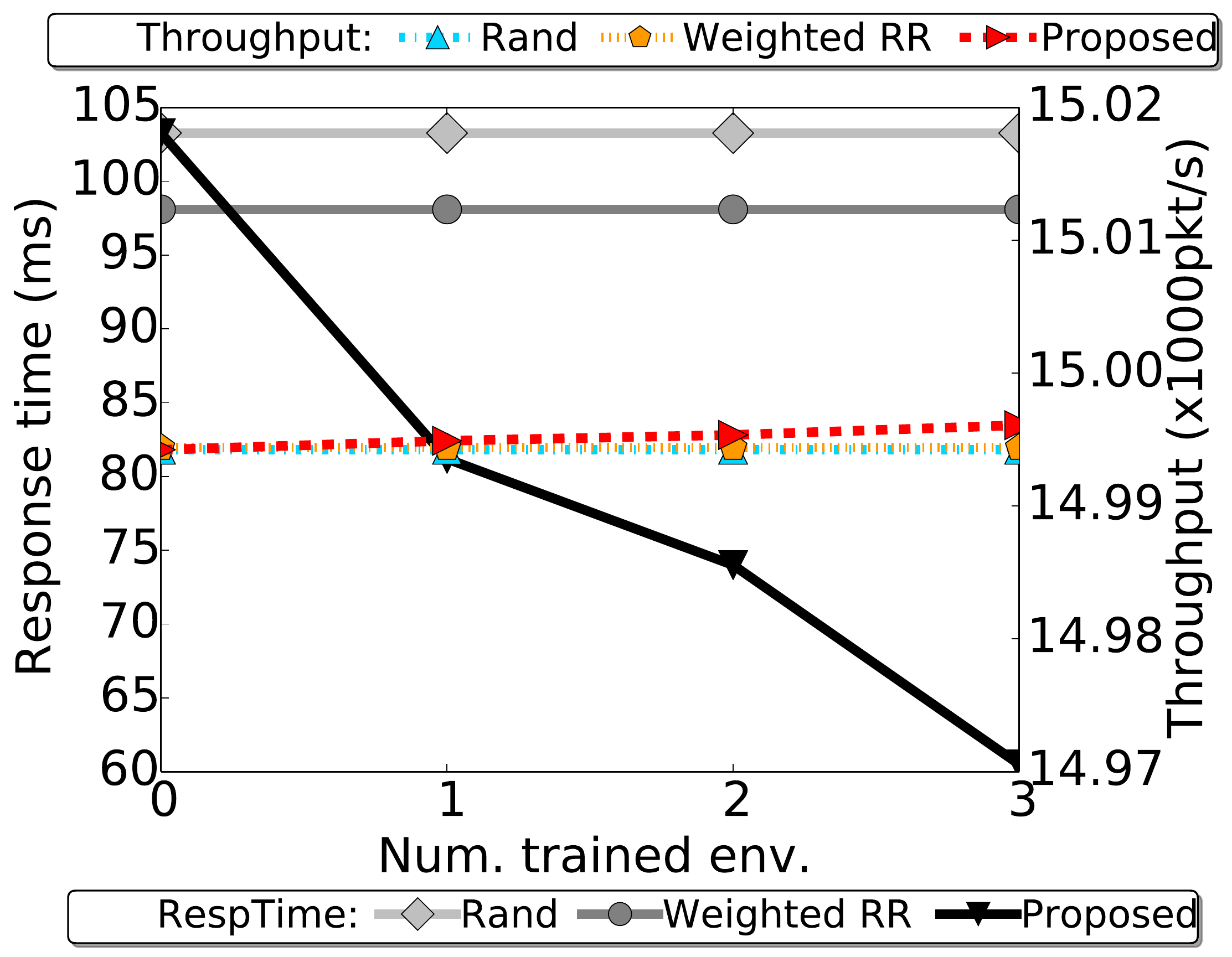}} \; 
    \subfloat[South American Network with arrival rate $20000$pkt/s (Env 5).]{\label{sec5_fig:4ctl_9sw}\includegraphics[width=.46\linewidth]{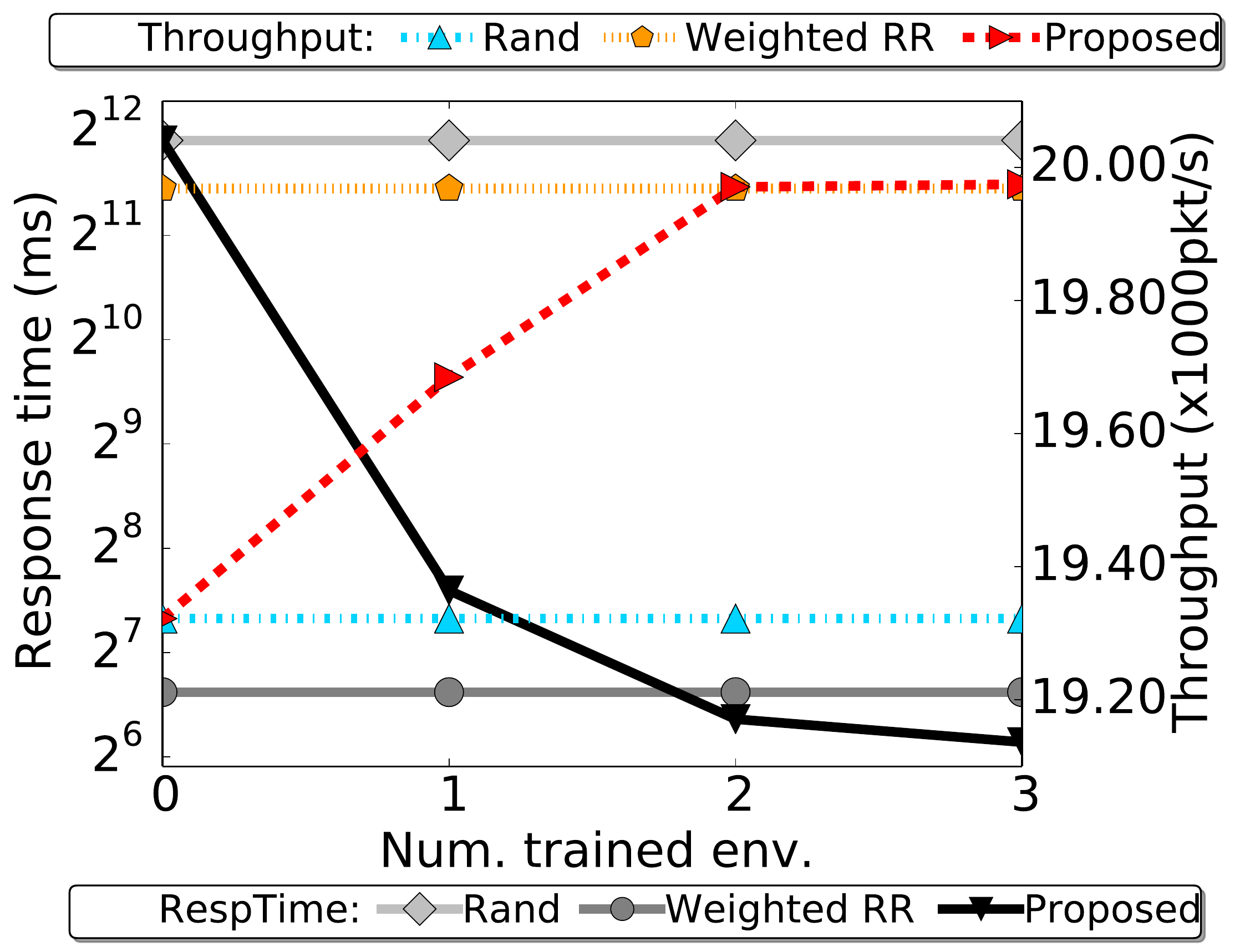}} \; \\
    \subfloat[Asian Network with arrival rate $15000$pkt/s (Env 6).]{\label{sec5_fig:4ctl_6sw}\includegraphics[width=.46\linewidth]{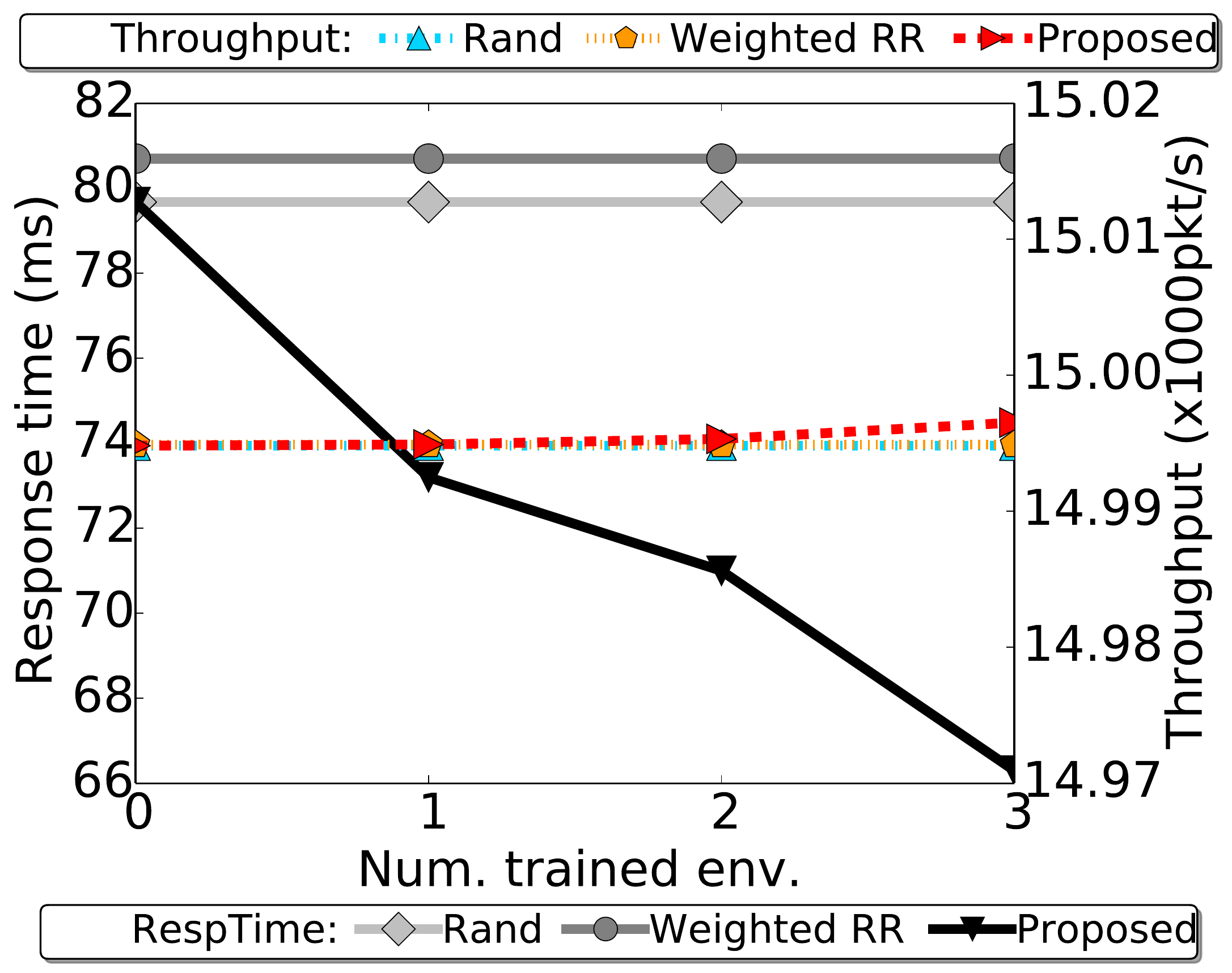}} \; 
    \subfloat[Asian Network with arrival rate $25000$pkt/s (Env 7).]{\label{sec5_fig:4ctl_12sw}\includegraphics[width=.46\linewidth]{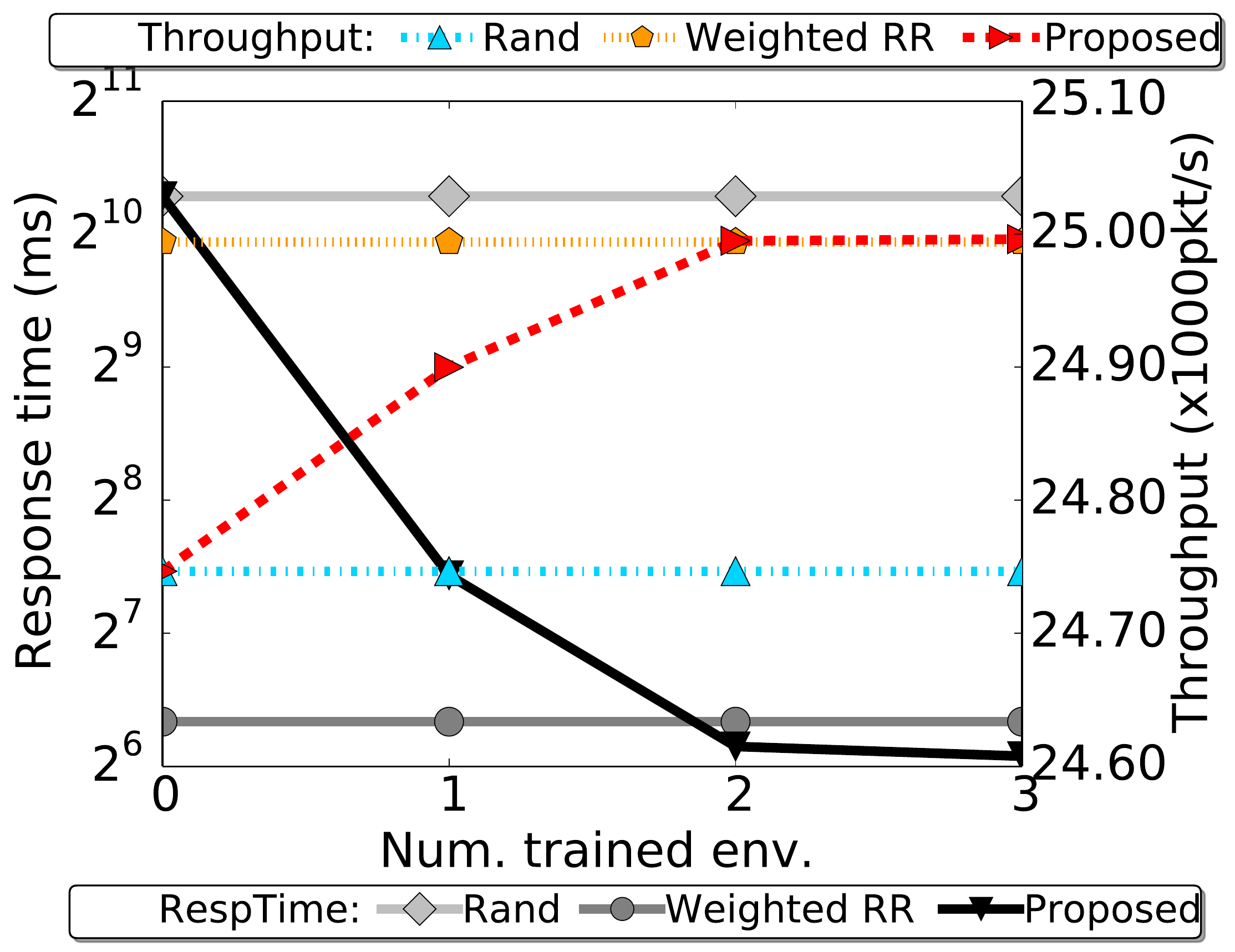}}
  \end{center}
  \caption{Testing performance comparison of random heuristic (Rand), weighted round-robin heuristic (Weighted RR), and our proposed trained SF (Proposed) on $2$ regional Sprint Network.
  The performance improvement of our proposed SF is shown by using the SF obtained at different training stages.}
\label{sec5_fig:testing_performance_4ctl}
\vspace{-5pt}
\end{figure}

\textbf{Testing Performance:} 
To demonstrate the effectiveness of our proposed approach, we compare the trained SFs obtained at different training stages with two competing scheduling heuristics: random (Rand) and weighted round-robin (Weighted RR) \cite{huang2017blac}. The topologies we use are South American (Env $4$ and Env $5$ in Table \ref{sec5_tab:network_setting}) and Asian Sprint networks (Env $6$ and $7$ in Table \ref{sec5_tab:network_setting}) \cite{sprint}.  
Due to various network sizes, the number of controllers deployed in each network is different as shown in Table \ref{sec5_tab:network_setting}. The locations of controllers are decided by k-center \cite{controllerPlacement}. Given the topology, the communication delay between any two nodes in the network is calculated using Dijkstra's algorithm \cite{dijkstra1959note}.  

The simulation begins with Env 4 where $3$ controllers are deployed. The total number of requests generated by the entire data plane is average to be $15000$pkt/s. It can be seen from Fig. \ref{sec5_fig:testing_performance_4ctl}(a) that the response time of our SF is initially similar to Rand. However, as the SF is trained with more network settings, its response time significantly drops from $105$ms to $60$ms, which is $40\%$ lower than both Rand and Weighted RR. Similar conclusions can also be drawn from Fig. \ref{sec5_fig:testing_performance_4ctl}(b)-(d) where different network settings are applied.
This is expected because our SF takes both the controller capacity and communication delay into account during request scheduling. On the other hand, Rand evenly schedules requests regardless of the communication delay or the controller capacity. Compared with Rand, Weighted RR distributes requests based on the controller capacity and achieves slightly better performance in general. However, in the network that spans large geographic areas, the communication delay contributes a significant proportion to the average response time. Solely considering the controller capacity obviously cannot achieve good performance. 

\begin{figure}[!tbp]
  \centering
  \includegraphics[width=.58\linewidth]{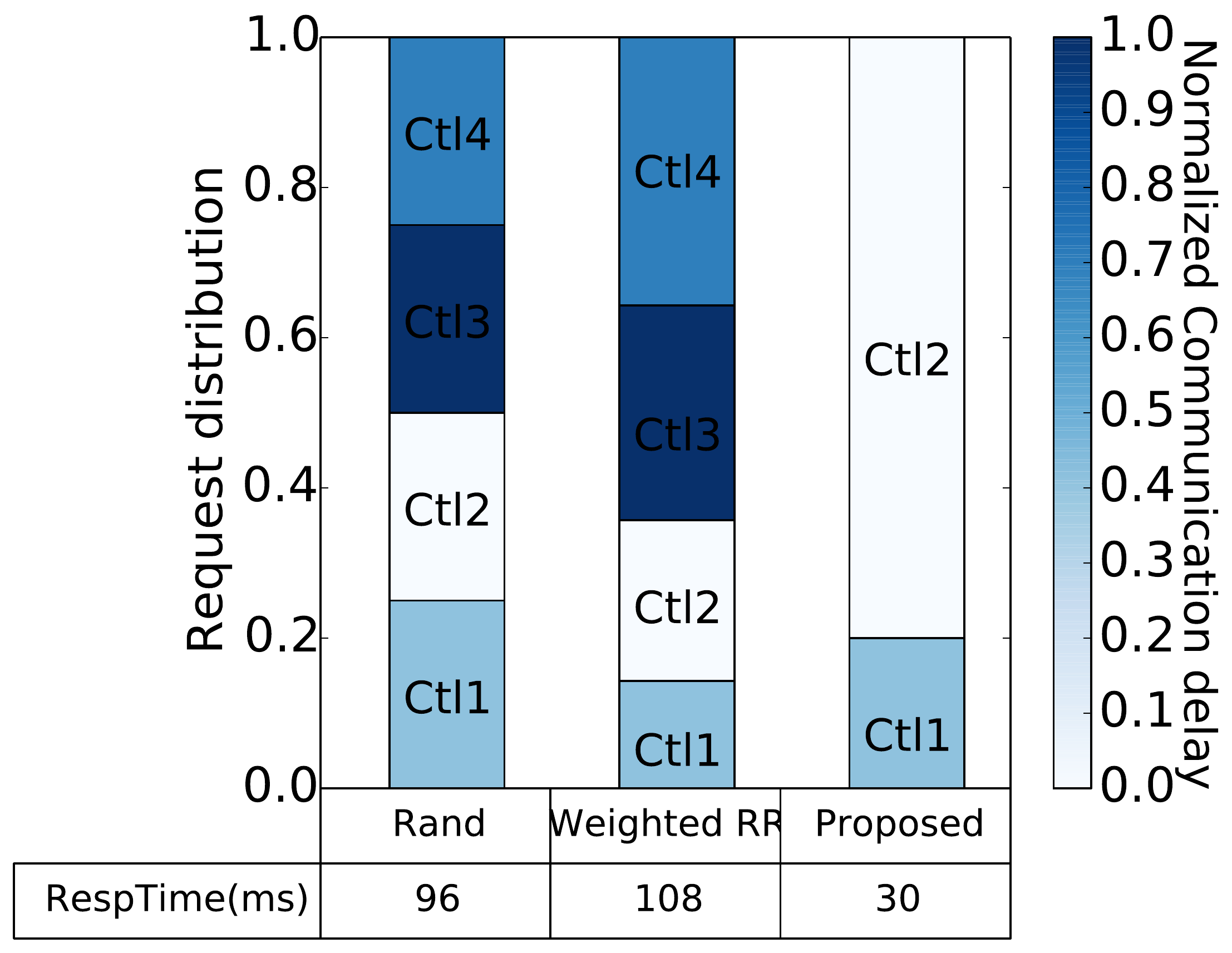}
  \caption{Request distribution over $4$ controllers obtained from a switch in Env $6$. For the proposed method, we use the SF obtained at the final training stage.}
  \label{sec5_fig:distribution}
  \vspace{-7pt}
\end{figure}

To verify this, we compare the request distributions of different approaches used in a switch in Env $6$. We can see from Fig. \ref{sec5_fig:distribution} that both Rand and Weighted RR schedule requests as we expected. On the other hand, instead of dispatching requests to all controllers as both Rand and Weighted RR do, our SF only schedules requests to controllers with low communication delay (i.e., Ctl1 and Ctl2) without overloading them, achieving the lowest response time ($30$ms). 

Another interesting phenomenon we can notice from Fig. \ref{sec5_fig:testing_performance_4ctl}(c) is that Rand outperforms Weighted RR on response time in Env $6$. The most likely reason is that the controller with smaller capacity is placed nearer to the switches compared to the larger-capacity one in the network. Thus, Rand sends more requests to nearby controllers, potentially reducing the response time.
Apart from that, all three approaches achieve similar throughput in Fig. \ref{sec5_fig:testing_performance_4ctl}(a) and Fig. \ref{sec5_fig:testing_performance_4ctl}(c), which is expected since the overall arrival rate is still below the whole network capacity and none of the controllers is overloaded.  

We also compare the performance of our trained SF with a different request arrival rate. As shown in Fig. \ref{sec5_fig:testing_performance_4ctl}(b), we notice that the throughput of Rand is $700$pkt/s smaller than Weighted RR due to controller overloading. In particular, the total arrival rate in Fig. \ref{sec5_fig:testing_performance_4ctl}(b) is $20000$pkt/s and each controller will evenly receive around $6666$pkt/s in Rand. Thus, the controller with $6000$pkt/s capacity will be inevitably overloaded, leading to the increase of response time and decrease of throughput. In comparison to Rand, the throughput of our trained SF increases as more network settings are trained, which effectively sends requests to near controllers without overloading them.  Similar phenomenon can be seen in Fig. \ref{sec5_fig:testing_performance_4ctl}(d).

Furthermore, it should be noted that, although our SF is trained in networks with two controllers, our testing results clearly show that it can generalize well to networks with more controllers. Therefore, our RL approach can design general and efficient SFs.

\section{Conclusions}
To effectively utilize the multi-controller resources in SDN, it is of great importance to design an SF that dispatches requests from switches to appropriate controllers. Motived by this, we propose an RL-based approach to solve the SFD problem by automatically learning an effective and generally applicable SF. Specifically, we formulate the SFD problem as an RL problem and a dispatching system is developed where an SF is in the form of an NN to calculate the priority of every controller.
After that, a specially designed selection scheme is applied to make the final dispatching decision using the obtained priorities. Along with the new dispatching system, a new training approach is developed which constantly improves the performance of the SF under different network settings via an adapted RL algorithm. Our simulation study showed that by using the newly proposed training approach, the SF can quickly converge to the optimal performance. Apart from that, the trained SF can generalize well and achieve significantly better performance compared with other heuristics.

\bibliographystyle{IEEEtranS}
\bibliography{version5_shrink}

\end{document}